\documentclass{article}
\usepackage{float}

\usepackage[final,dblblindworkshop]{neurips_2026}
\workshoptitle{Machine Learning and the Physical Sciences}
\makeatletter
\renewcommand{\@noticestring}{9th Workshop on Machine Learning and the Physical Sciences (ML4PS 2026).}
\makeatother

\ifdefined\LOCALFONTCHECK
  
\fi

\usepackage[utf8]{inputenc}
\usepackage[T1]{fontenc}
\usepackage{hyperref}
\hypersetup{pdftitle={Curvature Mismatch and Local Errors in Galaxy Property Estimation},
  pdfauthor={Austin Lutterbach, Angus Hoy, Jasper Chong, Aman Kumar, Mike Smith}}
\usepackage{url}
\usepackage{booktabs}
\usepackage{amsmath,amssymb}
\usepackage{microtype}
\usepackage{xcolor}
\usepackage{graphicx}

\newcommand{\Htan}{H^{S}}
\newcommand{\norm}[1]{\lVert #1 \rVert}

\title{Curvature Mismatch and Local Errors in Galaxy Property Estimation}

\author{%
  Austin Lutterbach\\
  EleutherAI SOAR\\
  UniverseTBD\\
  \texttt{rmhslutterbach@gmail.com}\\
  \And
  Angus Hoy\\
  EleutherAI SOAR\\
  UniverseTBD\\
  \texttt{angusjwh@gmail.com}\\
  \And
  Jasper Chong\\
  University of Queensland\\
  EleutherAI SOAR\\
  UniverseTBD\\
  \texttt{jasper.chong@uq.edu.au}\\
  \AND
  Aman Kumar\\
  UniverseTBD\\
  EleutherAI\\
  IUCAA\\
  \texttt{kumaramanasci@gmail.com}\\
  \And
  Michael J. Smith\\
  AstroAI\\
  Harvard-Smithsonian Center for Astrophysics\\
  UniverseTBD\\
  EleutherAI\\
  \texttt{mike@mjjsmith.com}\\
}

\begin{document}

\maketitle

\begin{abstract}
Linear probes are widely used to evaluate learned representations and infer physical quantities from unlabelled data. The probe's predictions are from an affine function in ambient space restricted to the underlying representation, and consequently, even a representation that has complete information needed to predict a target can produce systematic probe errors when that information is encoded in a form inaccessible to the linear probe. Here we share that some of this probing error is due to the innate geometry of the manifold. Linear probes, when evaluated on a curved manifold, generally have a nonzero intrinsic Hessian (what we colloquially call ``probe-facing curvature'') given by $\langle w_N,\mathrm{II}\rangle$, the manifold's second fundamental form dotted with the probe's normal component. We fit linear probes to galaxy embeddings from five pretrained encoders and intervene on this curvature term to evaluate its contribution to the probe performance. We find that keeping this inherited curvature improves the local fit where it matches the label's curvature. Removing the curvature component of the probe where it does not match the label curvature also provides improvements, and correcting it in every neighbourhood raises held-out local $R^2$ by 0.006 to 0.010 on galaxies, gaining more where the two curvatures disagree. We isolated these gains to be from the underlying geometry and not an automatic guarantee from least squares training.
\end{abstract}

\section{Introduction}

Linear probes evaluate learned representations by measuring how accurately targets can be recovered from frozen embeddings~\citep{alain2017probes}. Local reliability matters in physical science, where probes infer quantities without direct measurements. For galaxy images, linear probes predict catalogue quantities such as magnitude, redshift and stellar mass~\citep{smith2024astropt}, but aggregate scores alone do not characterize local prediction errors or explain their geometric origin~\citep{eyuboglu2022domino}. We show that the inherited curvature from the evaluated representation can impact the performance of the linear probe and captures a local source of prediction error.

We contribute (i) an implementation and synthetic validation of decoder-based second-fundamental-form estimation for local probe analysis; (ii) the application of probe-facing curvature $\operatorname{Hess}_M(w\cdot x)=\langle w_N,\mathrm{II}\rangle$ to a fitted probe, with a residual expansion identifying its second-order approximation error; (iii) evidence from 86{,}471 galaxies and five encoders that embedding curvature can positively impact the probe performance in ``lucky'' geometry-alignment scenarios and hinder it when misalignment occurs; and (iv) a bend correction that lets the probe's weight follow the manifold's bending in each neighbourhood, raising held-out local $R^2$ more where that bending is misaligned with the label. Code, records and notebooks are at \url{https://github.com/alutterb/galaxy-curvature}.

\section{Pointwise curvature from a decoder}
\label{sec:instr}

\paragraph{Data.} 86{,}471 galaxies from the AstroPT galaxy set~\citep{smith2024astropt} with catalogue labels ($r$-band magnitude, photometric redshift, smooth fraction, stellar mass), embedded by a supervised ImageNet-21k ViT-B/16 (``ViT-B'' below) and by four further encoders (Appendix~A); embeddings from the Platonic Universe release~\citep{duraphe2025platonic}. We unit-normalize embeddings (data on $S^{D-1}$; $D=768$ for ViT-B).

\paragraph{Instrument.} Let $F:\mathbb{R}^d\to\mathbb{R}^{D}$ be the decoder of a plain auto-encoder (three hidden layers of width 250, SiLU), fit on 80\% of rows; we evaluate curvature on held-out rows. We compute $J = DF(z)$, metric $g = J^{\top}J$, normal projector $P_N = I - Jg^{-1}J^{\top}$ and second fundamental form $\mathrm{II} = P_N D^2F(z)$, all by automatic differentiation \citep[cf.][]{lee2023curvature,acosta2023extrinsic} at points where $DF$ has full column rank $d$. We differentiate the projected decoder $F/\norm{F}$, whose image lies on the unit sphere, so that $\mathrm{II}$ splits exactly into the sphere's universal radial part and an in-sphere shape part (Eq.~\eqref{eq:split}).

\paragraph{Validation.} On a synthetic in-sphere surface at $D=768$ with closed-form $\mathrm{II}$ (Appendix~A), the decoder's estimate matches the true in-sphere trace $\Htan$ of $\mathrm{II}$ in direction (cosine 0.999), magnitude (ratio 1.00) and rank across points (0.94); the full tensor $K_S$ has median cosine 0.96 (Appendix~A).

\section{Setup and notation}
\label{sec:notation}

We assume the manifold hypothesis~\citep{fefferman2016testing}, that the embeddings lie near a smooth low-dimensional manifold. 

\paragraph{Notation.} Symbols used in the rest of the paper:
\begingroup\setlength{\leftmargini}{1em}
\begin{itemize}\setlength{\itemsep}{0pt}\setlength{\parskip}{0pt}
\item \emph{Manifold.} $M\subset S^{D-1}\subset\mathbb{R}^D$ is a smooth $d$-dimensional \textbf{manifold} ($d<D$) with \textbf{anchor} $x_0$, orthonormal \textbf{tangent frame} $J$ and \textbf{normal coordinates} $u$ (the chart), so $x(u)=x_0+Ju+\tfrac12\mathrm{II}(u,u)+O(\|u\|^3)$~\citep{lee2018riemannian}, where $\mathrm{II}$ is the normal-valued \textbf{second fundamental form}. On the sphere $\mathrm{II}=\mathrm{II}^S-g\,x$, with in-sphere \textbf{shape part} $\mathrm{II}^S\perp x$ and \textbf{trace} $\Htan$. Numerically $J=DF$, $g=J^\top J$ ($g=I$ in the orthonormal frame) and $u=g^{-1}J^\top(x-x_0)$.
\item \emph{Probe.} $\hat y=w\cdot x+b_0$ with \textbf{ridge penalty} $\alpha$ ($\alpha=100$ \textbf{pre-registered}, $\alpha^{*}$ \textbf{tuned}); $w=w_T+w_N$ (\textbf{tangent}, \textbf{normal}) and $w_N=w_S+w_{\mathrm{rad}}$, with $w_S\perp x$ and $w_{\mathrm{rad}}=(w\cdot x)\,x$.
\item \emph{Curvatures.} $H=\operatorname{Hess}_M y$ for a smooth \textbf{target} $y$; $K=\operatorname{Hess}_M\hat y=\langle w_N,\mathrm{II}\rangle$, the \textbf{probe's curvature} (Eq.~\eqref{eq:hess}), which splits as $K=K_S+K_{\mathrm{sph}}$ (Eq.~\eqref{eq:split}) into the \textbf{shape term} $K_S=\langle w_S,\mathrm{II}^S\rangle$, set by how $M$ bends inside the sphere, and the \textbf{sphere term} $K_{\mathrm{sph}}=-(\hat y-b_0)\,g$, which every probe on the sphere carries and which depends only on the prediction; the \textbf{mismatch} $\Delta=H-K$; $R=H-K_{\mathrm{sph}}$, the part of $H$ left for $K_S$ to reproduce: with $K_{\mathrm{sph}}$ fixed, $\Delta=R-K_S$; and the \textbf{gain} $\Gamma=2\langle H,K\rangle_g-\|K\|_g^2=\|H\|_g^2-\|\Delta\|_g^2$, positive when the probe's bending lowers the curvature part of local error.
\item \emph{Inner product.} $\langle A,B\rangle_g=g^{ia}g^{jb}A_{ij}B_{ab}$ and $\|A\|_g^2=\langle A,A\rangle_g$ for symmetric 2-tensors, the metric's version of the Frobenius inner product.
\item \emph{Patch and intervention.} On a \textbf{patch} $u\sim N(0,s^2I_d)$, $\bar c$ is the mean of $y-\hat y$ and $\delta=(dy-d\hat y)_J$ the \textbf{slope mismatch}. $\hat y_t=(w_T+w_{\mathrm{rad}})\cdot x+\tfrac t2K_S(u,u)+c$ scales $K_S$ by $t$ and refits $c$; over the neighbours, $q_c$ is the centred $\tfrac12K_S(u_i,u_i)$, $r_0$ the centred \textbf{residual} at $t=0$, and $t^{*}=\langle r_0,q_c\rangle/\|q_c\|^2$.
\end{itemize}
\endgroup

\section{Linear probe on a curved manifold}
\label{sec:theory}

With this setup, we know three things. First, although the
probe is affine in the ambient space, its
restriction to $M$ is generally nonlinear
in intrinsic coordinates:
\begin{equation}
(d\hat y)_J=J^\top w,\qquad
\operatorname{Hess}_M\hat y
=\langle w_N,\mathrm{II}\rangle=:K,
\label{eq:hess}
\end{equation}
where $K$ is the probe-facing curvature, a classical identity~\citep{milnor1963morse,chern1957total}.
Second, on the unit sphere it splits into
an in-sphere shape term and a radial term:
\begin{equation}
K=K_S+K_{\mathrm{sph}},\qquad K_S=\langle w_S,\mathrm{II}^S\rangle,\qquad K_{\mathrm{sph}}=-(\hat y-b_0)g.
\label{eq:split}
\end{equation}
Thus, the probe
inherits intrinsic curvature from the
geometry of the representation.

Third, on a patch of scale $s$, for sufficiently
smooth embedding and target functions
with controlled higher derivatives (Appendix~A),
\begin{equation}
\mathbb E_{\mathrm{patch}}[(y-\hat y)^2]=\bar c^{\,2}+s^2\|\delta\|^2+\tfrac12s^4\|\Delta\|_g^2+O(s^4\|\delta\|)+O(s^6).
\label{eq:resid}
\end{equation}
The Hessian mismatch therefore determines
the quadratic residual's contribution
to centred local error under this model.
Relative to the zero-Hessian baseline \(K=0\), the probe reduces the quadratic contribution to local error if and only if $\Gamma>0$. Hence $K$
must bend toward the target's
curvature, that is, lie closer to $H$ than $K=0$ does.
With $K_{\mathrm{sph}}$ fixed, the
corresponding condition for $K_S$
is
$2\langle R,K_S\rangle_g>\|K_S\|_g^2$.
Section~\ref{sec:real} tests this
relationship. Since the quadratic
contribution scales as $s^4$, our
empirical analysis controls for
variation in local neighbourhood scale.

We study the curvature term $\tfrac12 s^4\|\Delta\|_g^2$ because it is the only one that sees the full tensor $K$: $\delta$ is a first-derivative
mismatch, and the bending reaches $\bar{c}$ only through the trace of $\Delta$. 

\section{Experiments}
\label{sec:real}

We fit a five-fold out-of-fold ridge probe~\citep{alain2017probes} at the pre-registered $\alpha=100$ from embedding to label, score local $R^2$ inside each of 512 anchors' 2{,}048-neighbourhoods, and compute each curvature quantity's rank-partial Spearman against local $R^2$ with Freedman--Lane $p$-values~\citep{freedman1983permutation}, controlling for log patch radius, local label variance and count and, stronger, log radius at five scales. We also report the tuned $\alpha^{*}$ (Appendix~A). We use percentage ranges (x\%--y\%) to indicate the smallest and largest values observed across pairings of encoders and labels.

\begin{figure}[t]
\centering
\includegraphics[width=0.8\textwidth]{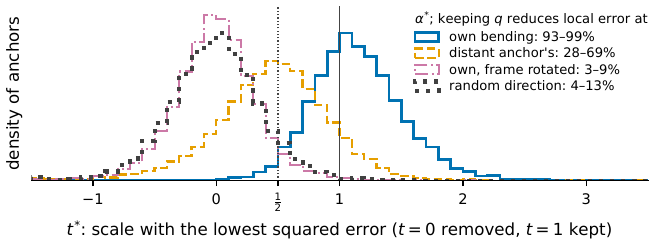}
\caption{The curvature the decoder measures at each galaxy is the curvature the fitted probe uses. With the probe's weight fixed, we score each anchor's 2{,}048 neighbours with its tangent and radial readout plus $t\,q$, for four quadratics $q$ of equal size. $t^{*}$ is the scale with the lowest squared error over those neighbours (the highest local $R^2$), and keeping $q$ fits better than removing it exactly when $t^{*}>\tfrac12$ (vertical dotted line). Own bending, $\tfrac12\langle w_S,\mathrm{II}^S\rangle$, peaks at $t^{*}=1$: the label-free decoder gives the bend at the size the probe uses. A distant anchor's $\mathrm{II}^S$ peaks near $\tfrac12$: the curvature is local. The own form with its tangent frame rotated (same principal curvatures) peaks at 0: its directions matter. Grey: a random normal direction, for reference. Tuned probe, five encoders at $d=16$, four labels.}
\label{fig:main}
\end{figure}

\paragraph{Probe curvature mismatch with label curvature.}
Across anchors a larger mismatch $\|\Delta\|_g$ goes with lower local $R^2$ for magnitude and redshift, but most of this is noise in the estimate of $H$; once it is removed only redshift keeps the association, and only at $\alpha=100$ (Appendix~B).
\paragraph{Keeping, removing or correcting the probe's bending.}
We then ask what the probe's bending does for its local fit. Holding the manifold and $(w_T+w_{\mathrm{rad}})\cdot x$ fixed, we score each anchor's 2{,}048 neighbours with $\hat y_t$ (Section~\ref{sec:notation}). So $t=1$ keeps the $K_S$ the probe inherits from the manifold and $t=0$ removes it.

Because $c$ is refit, the local error is a parabola in $t$: $\mathrm{SSE}(t)=\|r_0-t\,q_c\|^2$, minimised at $t^{*}$, and $t=0$ beats $t=1$ (i.e. an intervention) exactly when $2\langle r_0,q_c\rangle\leq\|q_c\|^2$. Across encoders and labels, $t=1$ beats $t=0$ at 79--99\% of anchors; with the tuned probe, that range increases to 93--100\%. 

We then ask if an intervention on curvature improves where misalignment occurs between the manifold and the label curvature. Since we cannot directly compare $\mathrm{II}$ and $H$, we compare $K_S$ with $R=H - K_{\text{sph}}$ to remove the sphere part and then calculate the cosine similarity between $K_S$ and $R$. At the tuned $\alpha^*$, an intervention $t=0$ improves over $t=1$ more often when there is misalignment compared to alignment. Among the top fifth misaligned regions of the manifold, intervention improves probe error on average 5.81\% of the time compared to 2.03\% of the time in the top fifth aligned areas. (See Appendix~C for full range of values).	

The tuned probe can still gain from a weight that changes from one neighbourhood to the next. In each neighbourhood we fit a bend correction on half of the neighbours, a small ridge fit that moves the tuned probe's weight along the decoder's 136 bending directions (from $\mathrm{II}^S$; Appendix~D) with the slope held fixed, and score it on the other half. Held-out local $R^2$ rises by 0.0059 to 0.0095 on ViT-B. Under half of that gain is curvature, and that part is larger where the probe's bend is less aligned with the label's curvature (rank-partial Spearman $-0.17$ to $-0.28$ for three of four labels). 

To separate the geometry from least squares (Figure~\ref{fig:main}), we keep the fitted weight $w_S$ and change only the bending it reads, rescaled to the same amplitude $\|q_c\|$. With the fit identical, least squares has no reason to prefer one bending over another. First, we take $\mathrm{II}^S$ from a distant anchor whose neighbourhood shares no galaxy with this one: the anchor's own bending beats it at 82--99\% of anchors. Second, we keep the anchor's own form but rotate its tangent directions, so the sheet bends by the same amounts along the wrong directions: it then helps at 17--34\% of anchors (3--9\% at $\alpha^{*}$), no more often than a random normal direction. So the gain depends on how the manifold bends at the anchor and along which directions, not only on the fitted weight.

\paragraph{Application to physical sciences.}
The curvature locates weak spots in a catalogue by neighbourhood. For catalogue work we suggest three rules. First, tune the ridge penalty. Areas where the manifold bending mismatches the label curvature requires the probe weight to compensate for the difference, and an overly aggressive penalty can prevent this correction. Second, remove the noise floor before reading a label's local curvature. Third, read accuracy by neighbourhood, not only globally. Grouping anchors by their neighbours' average $r$-band magnitude, the third with the dimmest galaxies has local $R^2$ up to 0.22 lower than the brightest third, for three of the four labels, and most of that drop is the labels' own noise, a larger share of their local variance there by 0.11 to 0.19 (Appendix~B). So low accuracy on faint galaxies mostly reflects the catalogue, not the probe: it calls for better labels or wider error bars there.

\section{Discussion}

A linear probe on a curved embedding inherits a bend from the manifold, and we have shown that this curvature can help the probe's performance when it aligns with the estimated label curvature and hinder it otherwise. We have also shown that when a probe is trained globally, the manifold fixes the probe's bend in each local neighbourhood, and when the label's curvature differs from that bend, it contributes to the probe's local error. 

These findings lead to three particular use cases: ensure the ridge penalty is tuned, as a strong penalty shrinks the probe's bend most and keeps it from matching the label's curvature; read accuracy by neighbourhood rather than globally, since the probe fits neighbourhoods of dim galaxies worst; and where a decoder is available, refine the probe's estimates locally with the bend correction.

\emph{Limitations.}
We study one galaxy sample, four targets and five encoders. About half of the estimated $\|H\|_g^2$ is noise floor (Appendix~B), and per-anchor residual curvature cannot rank single objects (split-half $\rho\approx0.33$). The intervention changes a local surrogate, not the encoder. Mismatch and alignment were defined after the pre-registered mean-curvature analysis and the intervention is exploratory; the permutation $p$-values ($^{*}$ in the tables) and the intervention's shares of anchors treat overlapping anchors as independent.

\section*{Acknowledgements}

We would like to thank Stella Biderman and EleutherAI for their compute and infrastructure support.

\bibliographystyle{unsrtnat}
{\small
\bibliography{references}
}

\newpage

\appendix
\section*{Appendix A: association results on ViT-B and across encoders}
\label{app:assoc}
\paragraph{Neighbourhoods and the label Hessian.} We draw the 512 anchors from the rows the decoder never saw; each anchor's $k=2{,}048$ neighbours are its nearest among all 86{,}471 galaxies by Euclidean distance between unit-normalized embeddings, that is, by cosine similarity. We estimate $H=\operatorname{Hess}_M y$ at each anchor by fitting a least-squares quadratic to the neighbours' labels in the tangent-projected coordinates $u=g^{-1}J^\top(x-x_0)$, whose Christoffel symbols vanish at the anchor, so the quadratic coefficient estimates the covariant Hessian in the small-neighbourhood limit.

\paragraph{Validation on a known surface.} At $D=768$ the in-sphere surface $G(z) = Q\,\mathrm{normalize}([\phi(z);\allowbreak\,0.8\,h(z);\allowbreak\,0\ldots])$ ($\phi$ inverse stereographic $\mathbb{R}^{16}\to S^{16}$, $h$ four Gaussian bumps, $Q$ a rotation) has closed-form curvature. The decoder's in-sphere trace $\Htan$ of $\mathrm{II}$ gives rank 0.94 against the true $\norm{\Htan}$, direction cosine 0.999 and magnitude ratio 1.00. With six labels of known Hessian on the same surface ($d=16$, $n=86{,}471$), the median tensor cosine of the decoder's $K_S$ with the truth is 0.963--0.965 without noise and 0.67--0.84 with isotropic off-surface noise whose size is a quarter of the median patch radius. Mean curvature, the scalar usually reported, keeps only the trace of $\mathrm{II}$ and fails accordingly, since its association with readout accuracy changes sign with the sampling on the known surface and with the latent dimension on galaxies.

\paragraph{Penalties and encoders.}
Tables~\ref{tab:xenc} and~\ref{tab:floor} give the probe at the pre-registered
$\alpha=100$ and at the tuned $\alpha^{*}$, chosen
per encoder and label by nested five-fold
cross-validation (one-standard-error rule) before
any geometry was computed at it; $\alpha^{*}$ is 0.1 or 0.32 in every cell.
Table~\ref{tab:xenc} holds the ViT-B analysis at $d=16$
and repeats it with four further encoders from the Platonic Universe release:
DINOv3 ViT-B/16, CLIP ViT-B, ConvNeXt-B and
ViT-L (widths 768, 512, 1{,}024 and 1{,}024;
all unit-normalized), each with a decoder of the same architecture and training
protocol (seed 0).
For magnitude and redshift the mismatch is
negatively associated with local $R^2$
in all five encoders at both penalties; most of this association is the label-Hessian noise floor (Appendix~B).

\begin{table}[h]
\centering
\scriptsize
\setlength{\tabcolsep}{1.3pt}
\begin{tabular}{ll*{10}{c}@{\hspace{5pt}}cc}
\toprule
 & & \multicolumn{2}{c}{ViT-B} & \multicolumn{2}{c}{DINOv3} & \multicolumn{2}{c}{CLIP-B} & \multicolumn{2}{c}{ConvNeXt-B} & \multicolumn{2}{c}{ViT-L} & \multicolumn{2}{c}{$^{\dagger}$ count} \\
\cmidrule(lr){3-4}\cmidrule(lr){5-6}\cmidrule(lr){7-8}\cmidrule(lr){9-10}\cmidrule(lr){11-12}\cmidrule(lr){13-14}
label & quantity & 100 & $\alpha^{*}$ & 100 & $\alpha^{*}$ & 100 & $\alpha^{*}$ & 100 & $\alpha^{*}$ & 100 & $\alpha^{*}$ & 100 & $\alpha^{*}$ \\
\midrule
mag\_r & mismatch & $-0.30^{\dagger}$ & $-0.24$ & $-0.58^{\dagger}$ & $-0.58^{\dagger}$ & $-0.29$ & $-0.34^{\dagger}$ & $-0.33$ & $-0.16$ & $-0.45^{\dagger}$ & $-0.49^{\dagger}$ & 3/0 & 3/0 \\
 & alignment & $+0.35$ & $+0.34$ & $+0.53$ & $+0.54$ & $+0.32$ & $+0.53$ & $+0.21$ & $+0.31$ & $+0.13$ & $+0.33$ & & \\
 & sphere & $-0.26$ & $-0.23$ & $-0.23$ & $-0.31$ & $-0.40$ & $-0.26$ & $-0.24$ & $-0.34$ & $-0.08^{*}$ & $+0.17$ & & \\
\addlinespace[2pt]
photo\_z & mismatch & $-0.39^{\dagger}$ & $-0.45^{\dagger}$ & $-0.60^{\dagger}$ & $-0.62^{\dagger}$ & $-0.42^{\dagger}$ & $-0.41^{\dagger}$ & $-0.46^{\dagger}$ & $-0.34$ & $-0.44^{\dagger}$ & $-0.49^{\dagger}$ & 5/0 & 4/0 \\
 & alignment & $+0.26$ & $+0.25$ & $+0.33$ & $+0.39$ & $+0.06^{*}$ & $+0.09$ & $+0.22$ & $+0.50$ & $+0.14$ & $+0.41$ & & \\
 & sphere & $-0.33$ & $+0.04^{*}$ & $-0.21$ & $+0.01^{*}$ & $-0.29$ & $+0.21$ & $-0.46$ & $+0.03^{*}$ & $-0.44$ & $+0.12$ & & \\
\addlinespace[2pt]
smooth\_fraction & mismatch & $-0.09^{*}$ & $-0.13$ & $-0.29$ & $-0.38^{\dagger}$ & $-0.24$ & $-0.34$ & $-0.19$ & $-0.11$ & $+0.01^{*}$ & $-0.20$ & 0/0 & 1/0 \\
 & alignment & $+0.19$ & $+0.27$ & $-0.02^{*}$ & $+0.22$ & $+0.24$ & $+0.27$ & $+0.04^{*}$ & $+0.26$ & $+0.14$ & $+0.38$ & & \\
 & sphere & $-0.13$ & $-0.10$ & $-0.06^{*}$ & $-0.22$ & $+0.05^{*}$ & $-0.24$ & $-0.15$ & $-0.19$ & $-0.11$ & $-0.03^{*}$ & & \\
\addlinespace[2pt]
stellar\_mass & mismatch & $+0.00^{*}$ & $+0.03^{*}$ & $+0.04^{*}$ & $-0.09$ & $-0.12$ & $-0.03^{*}$ & $-0.03^{*}$ & $-0.12$ & $-0.02^{*}$ & $-0.19$ & 0/0 & 0/0 \\
 & alignment & $+0.01^{*}$ & $+0.12$ & $-0.05^{*}$ & $+0.07^{*}$ & $-0.01^{*}$ & $-0.06^{*}$ & $+0.07^{*}$ & $+0.33$ & $-0.03^{*}$ & $+0.14$ & & \\
 & sphere & $-0.35$ & $-0.01^{*}$ & $-0.26$ & $-0.11$ & $-0.38$ & $+0.06^{*}$ & $-0.40$ & $+0.02^{*}$ & $-0.38$ & $-0.04^{*}$ & & \\
\bottomrule
\end{tabular}
\caption{Across encoders, $d=16$, 512 anchors: rank-partial Spearman correlation with local $R^2$, under the
multi-scale density control, of the mismatch $\norm{\Delta}_g$, the alignment $\langle H,K_S\rangle_g/(\norm{H}_g\norm{K_S}_g)$ and the size $\norm{K_{\mathrm{sph}}}_g=\sqrt d\,|\hat y-b_0|$ of the sphere term. Each column pair gives the probe at the pre-registered $\alpha=100$, then at the tuned $\alpha^{*}$.
$^{*}$ Not significant at 0.05. $^{\dagger}$ For the mismatch, the 95\% percentile interval of a cluster bootstrap excludes zero (the anchors cut into 32 blocks by average-linkage clustering on neighbourhood overlap, 2{,}000 resamples). The last pair counts the encoders whose mismatch partial is negative ($-$), and positive ($+$), with $^{\dagger}$. Before any $\alpha^{*}$ mismatch was computed we fixed the bar for a claim at three negatives of five.}
\label{tab:xenc}
\end{table}

\paragraph{Weak ridge.} The plain associations do not need an underfitting probe: at $\alpha=1$ (ViT-B; global $R^2$ up from 0.51 to 0.66) magnitude and redshift keep sign and significance (mismatch $-0.25$ to $-0.50$, alignment $+0.26$ to $+0.47$), and the tuned $\alpha^{*}$ agrees (Table~\ref{tab:xenc}).

\paragraph{Derivation of Eq.~\eqref{eq:resid}.} In the orthonormal frame ($g=I$ at $x_0$) write the residual $e=y-\hat y$ along $M$ as $e(u)=e_0+\delta^\top u+\tfrac12u^\top\Delta u+\rho(u)$. The quadratic coefficient is $\Delta$ because normal coordinates have vanishing Christoffel symbols at $x_0$ and $\operatorname{Hess}_M\hat y=K$ (Eq.~\eqref{eq:hess}); the controlled higher derivatives give $\rho(u)=O(\norm{u}^3)$. Exactly, $\mathbb E[e^2]=\bar c^{\,2}+\operatorname{Var}(e)$. For $u\sim N(0,s^2I_d)$ the linear and quadratic parts are uncorrelated (odd moments vanish), $\operatorname{Var}(\delta^\top u)=s^2\norm{\delta}^2$ and $\operatorname{Var}(\tfrac12u^\top\Delta u)=\tfrac12s^4\norm{\Delta}^2$. The covariance of the linear part with the cubic part of $\rho$ is $O(s^4\norm{\delta})$, and every other term is $O(s^6)$, which gives Eq.~\eqref{eq:resid}.

\section*{Appendix B: the label-Hessian noise floor (additional analyses)}
\label{app:floor}
Since $\Delta$ is the Hessian of the residual $y-\hat y$, some association with local accuracy is expected by construction (Eq.~\eqref{eq:resid}), but much of it is fitting noise. The probe's $K$ barely enters it: over the 40 encoder, label and penalty cells at $d=16$, the partials of $\norm{\Delta}_g$ and of the label-Hessian norm $\norm{\hat H}_g$ differ by a median 0.016 and correlate at 0.988.

\paragraph{The floor.}
The label Hessian $\hat H$ is a least-squares quadratic on $n$ noisy labels. Conditional on the neighbours, let the label noise $\varepsilon$ have mean zero and diagonal covariance $\Sigma$, and let $A$ be the $n\times p$ local quadratic design, $C=(A^{\top}A)^{-1}$ with quadratic block $C_{hh}$, $W$ the Gram matrix of $\norm{\cdot}_g$ on quadratic coefficients and $\bar H$ the noise-free Hessian. The fitted quadratic coefficients are $\hat h=\bar h+[CA^{\top}\varepsilon]_h$, so $\mathbb E\hat H=\bar H$ but
\begin{equation}
\mathbb E\norm{\hat H}_g^2=\norm{\bar H}_g^2+\operatorname{tr}(WV),\qquad V=[CA^{\top}\Sigma AC]_{hh}
\label{eq:floor}
\end{equation}
($\sigma^2\operatorname{tr}(WC_{hh})$ for homoscedastic noise). For $K$ fixed given the neighbours, $\norm{\hat H-K}_g^2$ carries the same term and $\langle\hat H,K\rangle_g$ has none, so the gain $\Gamma=2\langle\hat H,K\rangle_g-\norm{K}_g^2$ is floor-free. The upward bias of a squared estimate and its removal by products across independent halves are standard~\citep{hall1987integrated,walther2016reliability}. We estimate the floor from the single fit's residuals with the HC2 sandwich estimator of~\citet{mackinnon1985some}, or cancel it with the cross-fit product $\langle\hat H_A,\hat H_B\rangle_g$ of fits on two random halves of the patch. The same noise lowers local accuracy. With $P_A$ the projector onto the columns of $A$, $\hat\sigma_q^2=\norm{(I-P_A)y}^2/n$ and $\mathrm{lv}$ the patch variance of $y$, any probe quadratic in the chart on the patch has $1-R^2\ge\hat\sigma_q^2/\mathrm{lv}$, with equality iff it equals the label's local quadratic fit, since $y-\hat y=(I-P_A)y+(P_Ay-\hat y)$ with the two parts orthogonal. Noisier labels thus raise $\norm{\hat H}$ and $\norm{\hat H-K}$ and lower $R^2$ before any geometry enters.

\paragraph{On galaxies.}
The cross-fit product puts the floor at 45--66\% of $\norm{\hat H}_g^2$ (the median over anchors of $1-\langle\hat H_A,\hat H_B\rangle_g/\norm{\hat H}_g^2$; median 0.51 over the 20 encoder-label cells), and the HC2 correction agrees with it (median ratio 0.95--1.04). $\hat\sigma_q^2$ is negatively associated with local $R^2$, interval excluding zero, in 40 of 40 cells, and the bound holds at 98--100\% of anchors at $\alpha=100$, where the probe is locally quadratic to within 1--4\% of its error (not at $\alpha^{*}$). With the floor removed (Table~\ref{tab:floor}), magnitude keeps its association in no encoder, and redshift keeps it in three of five at $\alpha=100$ and in at most two at $\alpha^{*}$. The cross-fit mismatch $\langle\hat H_A-K,\hat H_B-K\rangle_g$ reads the same way: magnitude zero of five at both probes, redshift four of five at $\alpha=100$ and two at $\alpha^{*}$. These nulls are the floor cancelling, not a loss of power, since the half-patch norm's partial is more negative than the full patch's in 38 of 40 cells. Both estimators count what the chart quadratic does not fit as floor.

Splitting the probe's local error exactly, $\norm{y-\hat y}^2=\norm{(I-P_A)y}^2+\norm{(I-P_A)\hat y}^2-2\langle(I-P_A)y,(I-P_A)\hat y\rangle+\norm{P_A(y-\hat y)}^2$, with the last term split by the constant, linear and quadratic blocks of $A$, gives at $\alpha=100$ median shares of 0.69--0.79 for the label's residual after its local quadratic, 0.01--0.04 for the probe's off-quadratic part, $-0.14$ to $-0.04$ for their cross term, and 0.01--0.05, 0.09--0.22 and 0.10--0.15 for the constant, linear and quadratic parts of the mismatch; the last is an upper bound, because the fitted label quadratic carries noise. Floor-free, $K$ captures a median 0.11 ($\alpha=100$) and 0.28 ($\alpha^{*}$) of the corrected $\norm{\hat H}_g^2$; we call this ratio, $\Gamma/(\norm{\hat H}_g^2-\widehat{\mathrm{floor}})$, the capture. It cannot reach 1 here, because it sets the decoder's $K$ against a Hessian fitted from data.

\begin{table}[h]
\centering\scriptsize\setlength{\tabcolsep}{1.3pt}
\begin{tabular}{ll*{10}{c}@{\hspace{5pt}}cc}
\toprule
 & & \multicolumn{2}{c}{ViT-B} & \multicolumn{2}{c}{DINOv3} & \multicolumn{2}{c}{CLIP-B} & \multicolumn{2}{c}{ConvNeXt-B} & \multicolumn{2}{c}{ViT-L} & \multicolumn{2}{c}{$^{\dagger}$ count} \\
\cmidrule(lr){3-4}\cmidrule(lr){5-6}\cmidrule(lr){7-8}\cmidrule(lr){9-10}\cmidrule(lr){11-12}\cmidrule(lr){13-14}
label & quantity & 100 & $\alpha^{*}$ & 100 & $\alpha^{*}$ & 100 & $\alpha^{*}$ & 100 & $\alpha^{*}$ & 100 & $\alpha^{*}$ & 100 & $\alpha^{*}$ \\
\midrule
mag\_r & $\norm{H}^2{-}\mathrm{floor}$ & $+0.06^{*}$ & $+0.16^{\dagger}$ & $-0.05^{*}$ & $+0.02^{*}$ & $-0.07^{*}$ & $+0.14$ & $-0.08^{*}$ & $+0.14$ & $-0.08^{*}$ & $+0.01^{*}$ & 0/0 & 0/1 \\
 & $\langle H_A,H_B\rangle$ & $+0.09$ & $+0.17^{\dagger}$ & $-0.06^{*}$ & $+0.02^{*}$ & $-0.05^{*}$ & $+0.13^{\dagger}$ & $-0.00^{*}$ & $+0.15^{\dagger}$ & $-0.08^{*}$ & $+0.01^{*}$ & 0/0 & 0/3 \\
 & split cos & \multicolumn{2}{c}{$+0.35$} & \multicolumn{2}{c}{$+0.35$} & \multicolumn{2}{c}{$+0.31$} & \multicolumn{2}{c}{$+0.29$} & \multicolumn{2}{c}{$+0.30$} & & \\
\addlinespace[2pt]
photo\_z & $\norm{H}^2{-}\mathrm{floor}$ & $-0.22$ & $-0.24$ & $-0.35^{\dagger}$ & $-0.36^{\dagger}$ & $-0.20^{\dagger}$ & $-0.16$ & $-0.29^{\dagger}$ & $-0.12$ & $-0.09$ & $-0.02^{*}$ & 3/0 & 1/0 \\
 & $\langle H_A,H_B\rangle$ & $-0.22^{\dagger}$ & $-0.26^{\dagger}$ & $-0.27^{\dagger}$ & $-0.27^{\dagger}$ & $-0.14$ & $-0.13$ & $-0.27^{\dagger}$ & $-0.12$ & $-0.06^{*}$ & $+0.02^{*}$ & 3/0 & 2/0 \\
 & split cos & \multicolumn{2}{c}{$+0.32$} & \multicolumn{2}{c}{$+0.24$} & \multicolumn{2}{c}{$+0.25$} & \multicolumn{2}{c}{$+0.30$} & \multicolumn{2}{c}{$+0.24$} & & \\
\addlinespace[2pt]
smooth\_fraction & $\norm{H}^2{-}\mathrm{floor}$ & $+0.14$ & $+0.22$ & $-0.16$ & $-0.13$ & $-0.11$ & $-0.05^{*}$ & $-0.14$ & $+0.06^{*}$ & $+0.26$ & $+0.36$ & 0/0 & 0/0 \\
 & $\langle H_A,H_B\rangle$ & $+0.11$ & $+0.20$ & $-0.13$ & $-0.09^{*}$ & $-0.06^{*}$ & $-0.02^{*}$ & $-0.13$ & $+0.04^{*}$ & $+0.21$ & $+0.28$ & 0/0 & 0/0 \\
 & split cos & \multicolumn{2}{c}{$+0.33$} & \multicolumn{2}{c}{$+0.31$} & \multicolumn{2}{c}{$+0.28$} & \multicolumn{2}{c}{$+0.32$} & \multicolumn{2}{c}{$+0.30$} & & \\
\addlinespace[2pt]
stellar\_mass & $\norm{H}^2{-}\mathrm{floor}$ & $+0.13$ & $+0.25^{\dagger}$ & $-0.00^{*}$ & $-0.03^{*}$ & $-0.05^{*}$ & $+0.09$ & $+0.03^{*}$ & $+0.08^{*}$ & $+0.05^{*}$ & $+0.02^{*}$ & 0/0 & 0/1 \\
 & $\langle H_A,H_B\rangle$ & $+0.14^{\dagger}$ & $+0.22^{\dagger}$ & $+0.04^{*}$ & $+0.01^{*}$ & $-0.02^{*}$ & $+0.11^{\dagger}$ & $-0.01^{*}$ & $+0.08^{*}$ & $+0.08^{*}$ & $+0.04^{*}$ & 0/1 & 0/2 \\
 & split cos & \multicolumn{2}{c}{$+0.31$} & \multicolumn{2}{c}{$+0.22$} & \multicolumn{2}{c}{$+0.19$} & \multicolumn{2}{c}{$+0.29$} & \multicolumn{2}{c}{$+0.21$} & & \\
\bottomrule
\end{tabular}
\caption{Association with local $R^2$ once the noise floor is removed, $d=16$, 512 anchors, multi-scale density control: rank-partial Spearman of the label-Hessian norm less its HC2 floor, $\norm{H}_g^2-\widehat{\mathrm{floor}}$, and of the cross-fit product $\langle H_A,H_B\rangle_g$ of the two half-patch fits. Split cos is the median cosine between $H_A$ and $H_B$; it does not depend on the probe. Column pairs, marks, counts and the bar as in Table~\ref{tab:xenc}.}
\label{tab:floor}
\end{table}

\paragraph{The linear floor on ViT-B.}
On ViT-B at $d=16$ we split the held-out error of each probe into what refitting an affine readout removes, what a flexible residual readout (near-OLS plus a kernel ridge fitted to its residual) removes beyond it, and the rest. Refitting removes 0.303--0.323 of the error of the $\alpha=100$ probe and $-0.000$ to $+0.001$ of the error of a probe tuned by leave-one-out cross-validation ($\alpha=0.1$ or $0.001$ by label, not the nested $\alpha^{*}$ of Appendix~A). The flexible readout removes a further 0.124--0.179 of the tuned probe's error, an estimate of the linear floor, not a bound. A pre-registered test asked whether the probe's curvature predicts where that floor is largest: over a rank model on local label variance, log patch radius, count and density, the whitened $\norm{K}_g$ of the best affine readout adds $\Delta\rho=-0.011$ $[-0.032,+0.011]$ for magnitude (no added value) and $-0.002$ $[-0.078,+0.025]$ for redshift (inconclusive); the intervals run from the 1.25th to the 98.75th percentile of a 32-block bootstrap.

\paragraph{Misalignment, held out.} On ViT-B we measure on one half of each neighbourhood the cosine between $R$ and $K_S$, with the noise floor of $\|R\|_g^2$ removed, and ask on the other half whether $t=0$ wins. Among the fifth of anchors with the lowest cosine, removal wins more often than among the fifth with the highest: by $+0.066$ at $\alpha=100$ (interval $-0.001$ to $+0.122$) and $+0.038$ at $\alpha^{*}$ ($+0.010$ to $+0.070$), pooled over labels; for magnitude by $+0.070$ and $+0.100$ (at $\alpha^{*}$ the only label whose interval excludes zero). A random bend of the same size shows no such contrast ($-0.025$ and $-0.005$).

\paragraph{Neighbourhood types.} For each ViT-B anchor we average the catalogue's magnitude, redshift and smooth fraction over its neighbours and compare the top and bottom thirds of anchors, post hoc. The neighbourhoods are mixed, so the thirds differ little: the cuts sit at 18.16 and 18.37 in magnitude, 0.180 and 0.206 in redshift, and 0.724 and 0.786 in smooth fraction. The probe captures more of magnitude's curvature in fainter neighbourhoods ($+0.104$ at $\alpha=100$), more of magnitude's, smooth fraction's and stellar mass's in smoother ones ($+0.067$ to $+0.098$ at $\alpha=100$), and less of redshift's at high redshift ($-0.036$ at $\alpha=100$, $-0.057$ at $\alpha^{*}$). Fainter and higher-redshift neighbourhoods are also noisier, and with 24 comparisons per quantity one or two intervals could exclude zero by chance. The probe fits faint neighbourhoods worst. In the faintest third its local $R^2$ is lower than in the brightest third by 0.13 for magnitude, 0.14 for redshift and 0.22 for smooth fraction at $\alpha=100$, and by 0.07, 0.10 and 0.22 at $\alpha^{*}$ (every interval excludes zero except magnitude's at $\alpha^{*}$); stellar mass does not follow ($+0.04$ and $+0.00$). The labels are noisier there relative to their spread: the residual share $\hat\sigma_q^2/\mathrm{lv}$, read in the 16 leading principal components of each neighbour cloud, without the decoder, is higher by 0.19, 0.11 and 0.18, and by the bound that alone lowers local $R^2$ at $\alpha=100$. The bending strength $\norm{\mathrm{II}^S}$, the norm of the in-sphere second fundamental form at the anchor, needs no label. It does not change with the neighbourhood's magnitude or redshift (rank correlation $-0.03$ and $-0.06$ across anchors) and is larger in smoother neighbourhoods ($+0.30$, interval $+0.05$ to $+0.51$).

\section*{Appendix C: removing the bend where the curvatures disagree}
\label{app:misalign}
Section~\ref{sec:real} compares how often removing the bend wins in the fifth of anchors where $R$ and $K_S$ are least aligned against the fifth where they are most aligned. On ViT-B at $\alpha^{*}$ we rank anchors by that cosine on one half of each neighbourhood, as in Appendix~B, and score the other half. Removal wins at an anchor when $t=0$ gives the lower squared error there, and its gain is $R^2(t{=}0)-R^2(t{=}1)$. Table~\ref{tab:misalign-r2} gives, for each fifth, the number of anchors where removal wins and how much it gains. Removal wins more often in the misaligned fifth for three of the four labels, but its gains are small: a median of 0.5 to 1.6 points of $R^2$ (6.9 for redshift, from two anchors). This analysis is post hoc.

\begin{table}[h]
\centering
\footnotesize
\begin{tabular}{lcccc}
\toprule
 & \multicolumn{2}{c}{aligned fifth} & \multicolumn{2}{c}{misaligned fifth} \\
\cmidrule(lr){2-3}\cmidrule(lr){4-5}
label & removal wins & $R^2$ gain & removal wins & $R^2$ gain \\
\midrule
mag\_r & 1 of 100 & $+3.2$ & 11 of 100 & $+1.6$ (12.4) \\
photo\_z & 0 of 99 & -- & 2 of 99 & $+6.9$ (13.6) \\
smooth\_fraction & 4 of 99 & $+0.8$ (1.3) & 3 of 99 & $+1.3$ (1.4) \\
stellar\_mass & 3 of 97 & $+0.9$ (1.6) & 7 of 97 & $+0.5$ (0.7) \\
\bottomrule
\end{tabular}
\caption{ViT-B, tuned probe ($\alpha^{*}$), 512 anchors; post hoc. Fifths of anchors by the cosine of $R$ and $K_S$ on half A of each neighbourhood: aligned (highest cosine) and misaligned (lowest). Removal wins: anchors where $t=0$ fits the held-out half B better than $t=1$. $R^2$ gain: median of $R^2(t{=}0)-R^2(t{=}1)$ over those anchors, in points of local $R^2$, with the largest in parentheses where there are two or more wins.}
\label{tab:misalign-r2}
\end{table}

\section*{Appendix D: the bend correction on molecular embeddings (additional analyses)}
\label{app:qm9}
A question we had after viewing the results on galaxy data was if the approach generalizes to other mediums. We repeated the bend correction of Section~\ref{sec:real} on QM9~\citep{ramakrishnan2014qm9}: 130{,}744 molecules embedded by MoLFormer-XL~\citep{ross2022molformer}, with four quantum-chemistry labels (the HOMO--LUMO gap, the dipole moment $\mu$, the isotropic polarisability and the heat capacity $c_v$). The setup matches ViT-B: a seed-0 decoder at $d=16$, 512 anchors with 2{,}048 neighbours each, and the probe at its tuned penalty. In each neighbourhood we refit the probe's curvature on one half of the neighbours and score it on the other half. Table~\ref{tab:qm9} sets the two datasets side by side.

\paragraph{The bend correction.} At $d=16$ a quadratic shape in the chart has 136 independent components, and the decoder's $\mathrm{II}^S$ at the anchor sends each to a direction in the embedding. We take its 136 singular directions $n_k$: moving the probe's weight along $n_k$ bends the probe in the $k$th shape, and a neighbour's coordinate along it is $n_k\cdot(x-x_0)$. On one half of the neighbours we fit a ridge correction to the tuned probe's residual on these 136 coordinates, after removing the part of them that is constant or linear in the chart, so the correction changes the probe's curvature and not its slope. Its penalty is chosen by leave-one-out on that half, with no correction as one choice, and we score the corrected probe on the other half. The corrected probe stays linear in the embedding, with a weight that changes from one neighbourhood to the next.

\begin{table}[h]
\centering\footnotesize
\begin{tabular}{lcc}
\toprule
 & QM9, MoLFormer-XL & galaxies, ViT-B \\
\midrule
curved share of a neighbourhood's held-out variance & 0.12--0.15 & 0.06--0.08 \\
curvature the tuned probe leaves unread (read low) & 0.025--0.080 & 0.006--0.016 \\
gain of a free quadratic in the chart & 0.023--0.041 & 0.0015--0.0039 \\
gain of the bend correction & 0.044--0.049 & 0.0059--0.0095 \\
\bottomrule
\end{tabular}
\caption{The bend correction of Section~\ref{sec:real} on two datasets, tuned probe, $d=16$, 512 anchors. Shares and gains are in held-out local $R^2$; each range runs over the four labels.}
\label{tab:qm9}
\end{table}

The bend correction removes 18 to 22\% of the tuned probe's local error for the gap, the polarisability and $c_v$, and 8\% for $\mu$, whose probe fits least well (local $R^2$ 0.35, against 0.74 to 0.80 for the others). The tuned probe reads about the same share of each label's curvature on both datasets: 0.40 to 0.69 here, 0.58 to 0.64 on ViT-B. This share is the regression coefficient of the probe's fitted chart Hessian on the label's, both fitted from the data on two halves of each neighbourhood, multiplied across the halves and pooled over anchors. So it does not carry the decoder-to-data gap of the ratio in Appendix~B. The molecular labels have about twice as much curvature to read, and the part the probe misses is three to five times larger, and a refit on half a neighbourhood recovers more of it (0.56 to 0.99 of the unread curvature here, against 0.26 to 0.33 on ViT-B). Our estimator reads these unread amounts low on both datasets. For each label's own linear fit, which the probe reproduces, it reports $-0.11$ to $-0.15$ of the curvature unread here and $-0.21$ to $-0.26$ on ViT-B, where the truth is near 0. So the unread row of Table~\ref{tab:qm9} and the ratios built on it are rough. The gains are differences between two fits and do not carry this bias.

Two limits apply. The bend correction ran on one encoder and one decoder. And here the decoder misses some structure. Held out, a free quadratic in the chart explains 0.19 to 0.36 of each normal coordinate, while the decoder's own bending explains 0.005 to 0.16; on ViT-B the decoder explained more than the free quadratic. So part of the molecular gain may come from how the decoder fits at $d=16$, above the encoder's estimated intrinsic dimension of 11. The bend correction's gain is also about half curvature (0.020 to 0.029) and half structure off the chart (0.019 to 0.026), and its curved part reaches 0.70 to 0.88 of what the free quadratic recovers, where on ViT-B it matched it. On ViT-B the curved part is 0.0022 to 0.0039 of the 0.0059 to 0.0095, and the rest is structure off the chart.

\end{document}